\documentclass[11pt,a4paper]{article}
\usepackage[hyperref]{emnlp2020}
\usepackage{times}
\usepackage{latexsym}

\usepackage{graphicx}
\usepackage{float}
\usepackage{amsmath}
\usepackage{amssymb}
\usepackage{multirow}
\usepackage{booktabs}
\usepackage{url}
\usepackage{array}
\usepackage{enumitem}
\usepackage{algorithm}
\usepackage{algorithmic}
\usepackage{pifont}
\usepackage{CJKutf8}
\usepackage{makecell}
\usepackage{stackengine}

\usepackage{microtype}

\aclfinalcopy 
\def\aclpaperid{} 

\title{Unsupervised Explanation Generation for Machine Reading Comprehension}

\author{Yiming Cui$^{1,2}$, Ting Liu$^1$, Shijin Wang$^{2,3}$, Guoping Hu$^2$, \\ 
{$^1$Research Center for SCIR, Harbin Institute of Technology, Harbin, China} \\
{$^2$State Key Laboratory of Cognitive Intelligence, iFLYTEK Research, China} \\
{$^3$iFLYTEK AI Research (Hebei), Langfang, China} \\
{$^1$\tt \{ymcui,tliu\}@ir.hit.edu.cn}\\
{$^{2,3}$\tt\{ymcui,sjwang3,gphu\}@iflytek.com}\\  
}

\date{}

\begin{document}
\begin{CJK*}{UTF8}{gbsn}
\maketitle

\begin{abstract}
With the blooming of various Pre-trained Language Models (PLMs), Machine Reading Comprehension (MRC) has embraced significant improvements on various benchmarks and even surpass human performances.
However, the existing works only target on the accuracy of the final predictions and neglect the importance of the explanations for the prediction, which is a big obstacle when utilizing these models in real-life applications to convince humans.
In this paper, we propose a self-explainable framework for the machine reading comprehension task.  
The main idea is that the proposed system tries to use less passage information and achieve similar results compared to the system that uses the whole passage, while the filtered passage will be used as explanations.
We carried out experiments on three multiple-choice MRC datasets, and found that the proposed system could achieve consistent improvements over baseline systems.
To evaluate the explainability, we compared our approach with the traditional attention mechanism in human evaluations and found that the proposed system has a notable advantage over the latter one.
\end{abstract}

\section{Introduction}
Machine Reading Comprehension (MRC) is a task that to read and comprehend passages and answer relevant questions, requiring a deep understanding of the passage and question as well as their relationships. 
With the latest pre-trained language models, some of the systems could surpass human performance.
While massive efforts have been made, most of the previous works were focusing on designing more sophisticated systems to achieve better evaluation scores and neglect the importance of the explainability of the system, which is crucial for the human to understand the decision process inside black-box artificial intelligent models. 

As a result, Explainable Artificial Intelligence (XAI) \citep{gunning2017explainable} received much more attention in these years than ever, which aims to produce more explainable machine learning models while preserving a high accuracy of the model output and let the humans understand its intrinsic mechanism.
To improve the explainability of the AI, one could seek decomposability of the traditional machine learning model (such as decision trees, rule-based systems, etc.), or use post-hoc techniques for deep learning models \citep{xai-survey}.

For the machine reading comprehension task, we could annotate additional explanations to the questions and let the model simultaneously learn to predict both the final answer and its explanation.
This is similar to HotpotQA \citep{yang-etal-2018-hotpotqa}, which is a multi-hop MRC dataset, where it requires the machine to predict both the final answer span and its supporting facts.
However, this requires a large amount of labeled data for the machine learning system to learn a reliable model, which is much time-consuming and expensive.

To both address the explainability of the machine reading comprehension systems and alleviate human annotations, in this paper, we propose a novel self-explainable mechanism called Recursive Dynamic Gating (RDG) to the existing models without relying on additional annotated data.
The RDG mechanism injects into the pre-trained language model and applies a filter in the input representations of each transformer layer to control the amount of the information.
Moreover, to improve the accuracy of the gating, we apply a convolutional operation to mimic the N-gram summarization of the local context.
In the inference stage, we extract the gating values to represent the selections of the input words as explanations. 
The contributions of this paper are listed as follows.
\begin{itemize}[leftmargin=*]
	\item To improve the explainability of the MRC model, we propose a self-explainable mechanism called Recursive Dynamic Gating (RDG) strategy to control the amount of information that flows into each transformer layer.
	\item The results on several multiple-choice machine reading comprehension datasets show that the proposed method could consistently improve objective benchmarks.
	\item Through careful human evaluation, the proposed method shows a notable advantage over the traditional attention mechanism in explainability.
\end{itemize}

\section{Related Work}\label{related-work}
Machine Reading Comprehension (MRC) is similar to the traditional question answering, but put more emphasis on the comprehension of the passage. 
Various datasets and models have been proposed \citep{dhingra-etal-2017,kadlec-etal-2016,cui-acl2017-aoa} since the emerge of the cloze-style reading comprehension \citep{hermann-etal-2015, hill-etal-2015}, which is regarded as the starting point of the recent MRC research.
Later, the release of SQuAD \citep{rajpurkar-etal-2016} becomes a milestone in MRC, which is a span-extraction reading comprehension task, and greatly accelerated the research in this area. 
With the blooming of the recent pre-trained language models, such as BERT \citep{devlin-etal-2019-bert}, XLNet \citep{yang2019xlnet}, RoBERTa \citep{liu2019roberta}, ALBERT \citep{Lan2019ALBERT}, etc., some of the systems could surpass human performance on several MRC datasets.

However, though the benchmarks have been refreshed to such a high level, the decision process and the explanations of these artifacts still remains unclear, raising concerns about the reliability of the artificial intelligence systems. 
In this context, Explainable Artificial Intelligence (XAI) becomes more important than ever. 
Conventional machine learning approaches, such as decision tree, could be decomposed into `explainable' components to reveal its explainability.
However, this is not the case for the recent machine learning approaches, especially in natural language processing, where the models are usually composed of deep neural networks, such as BERT. 
To better understand the internal mechanism of BERT-based architecture, \citet{kovaleva-etal-2019-revealing} discovered that, there are repetitive attention patterns across different heads in the multi-head attention mechanism indicating its over-parametrization. 

To achieve multi-hop and explainable question answering, HotpotQA \citep{yang-etal-2018-hotpotqa} is proposed, which requires the machine to not only give the final answer but also extract the evidence sentences in the passages. 
Following this dataset, various models have been proposed \citep{yang-etal-2018-hotpotqa,shao-2020-gnn-analysis} to improve the overall accuracy.
\citet{liu-etal-2019-towards-explainable} proposed to build a generative explanation framework for the text classification with two new datasets that contain summaries, rating scores, and fine-grained reasons.
However, most of these works rely on human annotations in order to do supervised learning, which will be a challenge in real-life applications.

Another direction is to seek a self-explainable system that does not require prior annotations for the explanations. 
For example, \citet{perez-etal-2019-finding} proposes to train an evidence agent to select the passage sentences that most convince the MRC model of a given answer using a subset of the passage.
\citet{moon-etal-2019-memory} propose a Memory Graph Network to enable dynamic expansion of memory slots through graph traversals to improve explainability in QA reasoning.
\citet{wang-etal-2020-evidence} applied distant supervision to generate pseudo evidence sentence labels and refined them through a deep probabilistic logic learning framework.

In this paper, we focus on designing a self-explanatory system for the machine reading comprehension task.
Inspired by Information Bottleneck (IB) principle \citep{tishby2000information}, we propose a novel mechanism called Recursive Dynamic Gating (RDG) to achieve better explainability for the BERT-like pre-trained language models without any additional annotations. 
Unlike previous works, our model aims to achieve better explainability of the answer while preserving or even surpass the original system in the objective evaluations. 
Instead of using a neural network model to filter a subset of the sentences in the passage, which could be non-differentiable and requires additional makeup (such as Gumbel-Softmax reparameterization \citep{Jang2017CategoricalRW}), we seek to control the amount of passage information in the hidden representations directly to avoid such circumstances.

\begin{figure*}[tbp]
  \centering
  \includegraphics[width=0.9\textwidth]{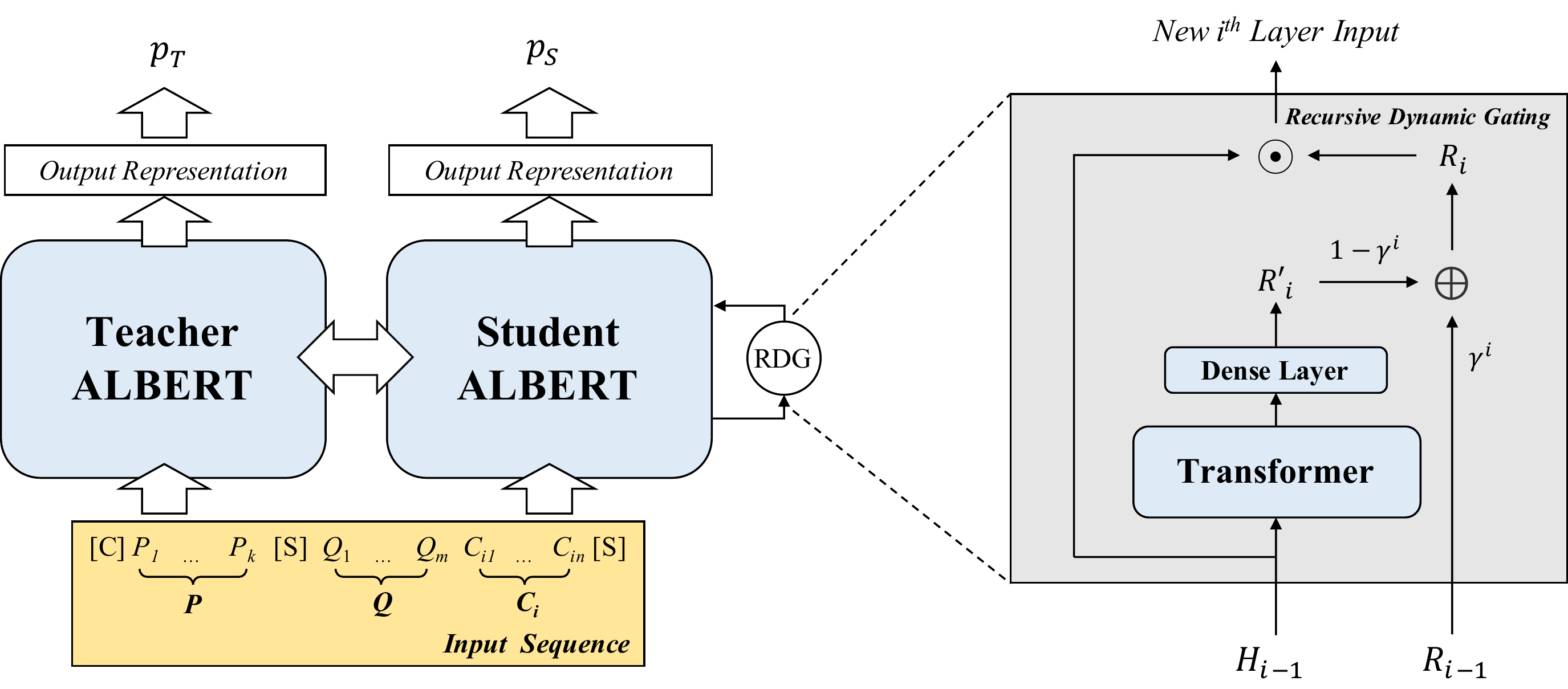}
  \caption{\label{model} System overview of the Recursive Dynamic Gating (RDG) mechanism for machine reading comprehension task. Note that, minor operations are omitted for simplicity.}
\end{figure*}

Unlike the attention mechanism \citep{bahdanau2014neural} in traditional non-transformer models, the attention information will be separated into several subsets in multi-head attention \citep{vaswani2017attention}, which is the key component in transformer-based models.
Though the multi-head attention has proved to be effective, it is harder for the human to get a comprehensive view of which part of the passage that the model attended to.
As a remedy, our RDG mechanism provides a straightforward explanation of the model prediction, which could be more friendly in the human view.

\section{The Approach}\label{rdg}
\subsection{Overview}
In this paper, we focus on the multiple-choice machine reading comprehension (MC-MRC) task, which requires the machine to read a passage and choose a correct answer among several candidates for the given question.
Unlike SQuAD \citep{rajpurkar-etal-2016}, the answer for the MC-MRC task is not an exact span in the passage, where the human needs to know the explanations for the answer predictions. 

We will overview our model on a general basis, to better understand its mechanism. 
The overall neural architecture of our model is depicted in Figure \ref{model}. 
The primary idea is to train a two-pass system, including the original fine-tuning system (teacher) and an explanatory system (student), which are jointly trained.
\begin{itemize}[leftmargin=*]
	\item {\bf Teacher}: uses the original input to learn the relations between the passage, question, and candidate answers.
	\item {\bf Student}: utilizes the RDG mechanism to filter the passage information in the spirit of using less passage information to achieve similar results to the teacher system.
\end{itemize}

In the inference stage, we use the teacher system to predict the final answer and use the student system to generate explanations.

\subsection{Formulation for ALBERT}
In this paper, we use ALBERT \citep{Lan2019ALBERT} as the backbone of our model for its effectiveness in machine reading comprehension as well as other NLP tasks. 
We omit rather extensive full formulations for ALBERT and only shows the high-level architecture.
Given an input sequence $X=\{x_1,..,x_l\}$ ($l$ is the maximum sequence length), ALBERT model will convert it into a contextualized representations $H_L\in\mathbb{R}^{l*h}$ through an embedding layer (which consists of word embedding, positional embedding, and token type embedding), and a consecutive $L$-layer transformer. 
Note that the weight for each transformer layer is identical through cross-layer parameter sharing.
Also, the teacher and student ALBERT model shares the same parameters with all transformer layers initialized by the pre-trained ALBERT weight.
\begin{gather} 
H_0 = \mathbf{Embedding}(X)  \\
H_i = \mathbf{Transformer}(H_{i-1}), i=\{1,...,L\}
\end{gather}

\subsection{Recursive Dynamic Gating}
Our recursive dynamic gating $R \in\mathbb{R}^{m}$ applies on the input representation of each transformer layer to dynamically control the information flow to the next layer, which mimics the process of multi-hop reading strategy.
Specifically, for $i$th transformer layer, we rewrite the Equation 2 as follows. 
\begin{gather} 
H_i = \mathbf{Transformer}(H_{i-1} * R_{i})
\end{gather}

Given previous transformer layer output representation $H_{i-1}$, we will construct recursive dynamic gating $R_i$ for $i$th layer as follows.
We first feed $H_{i-1}$ into a 1-layer transformer and squeeze the output into a scalar with respect to each word, indicating their importance, which is similar to the attention mechanism.
Note that, the transformer model for gating mechanism is also initialized with pre-trained weights but does not share with other transformer layers in ALBERT model.
\begin{gather}
R'_i = {\bf w}^{T}\cdot \mathbf{Transformer}(H_{i-1})
\end{gather}

To let the resulting gating value aware of its adjacent words, we propose a smoothing technique based on the convolutional operation.
We use a convolutional filter to dynamically summarize the gating value in a fixed window. 
Following \citet{kim-2014}, we apply the 1D convolution operation, which is commonly used in natural language processing tasks.
To keep the sequence length unchanged, we apply the {\em same-length} convolution with a kernel size of $k$.
Inspired by \citet{cui-etal-2019-cru}, we apply a residual connection to `highlight' original word in place. 
To keep consistent with original transformer implementation, we apply Gaussian Error Linear Unit (GELU) \citep{hendrycks-2016-gelu} for activation, where we clipped activation output in the range of $[0,1]$.\footnote{While it is natural to use sigmoid or softmax function, we did not find additional gains by using these functions.}
\begin{gather}
\hat{R}_i = \mathbf{GELU}(conv(R'_i, k) + R'_i)
\end{gather}

As the gating mechanism will diminish the input representations (in extreme cases, all representation will be erased, i.e., multiplied by zeros), to avoid sudden severe changes, we apply a recursive learning approach. 
In this paper, we investigate two variants: recursive history (Eq. 6a) and constant history (Eq. 6b), where $I \in\mathbb{R}^{\mathcal \\l}$ represents the input mask (padding positions are 0 and others are 1). 
The final gating value will be determined by both historical gating value $R_{i-1}$ and the calculated current gating value $R'_i$, with a history decay coefficient $\gamma \in (0,1)$.
We observe that the recursive history (Eq. 6a) is better for the larger dataset as the gating mechanism will be learned much adequately, while the constant history (Eq. 6b) is suitable for smaller dataset.  
\begin{subequations}
\begin{gather}
\tilde{R}_i = \gamma^i*R_{i-1} + (1 - \gamma^i)*\hat{R}_i \\
\tilde{R}_i = \gamma^i*I + (1 - \gamma^i)*\hat{R}_i
\end{gather}
\end{subequations}

As our aim is to filter the passage, we apply a mask to the $\tilde{R}_i$ to preserve the gating values in the passage and set constant ones in the question and choice to protect their representations from diminishing. 
In practice, this could be achieved by the calculations between the input mask $I$ and token type ids $T$ (where the positions of the question and choices are set to 1, and others are 0).
\begin{gather}
R_i =  \tilde{R}_i * I * (1 - T) + T
\end{gather}

In this way, we've injected recursive dynamic gating into the ALBERT model.
Finally, we apply a linear transformation ${\bf w_o} \in\mathbb{R}^{\mathcal \\h}$ to get final logits for both teacher and student model, as shown below, where $H_T$ and $H_S$ represents their final output of ALBERT model.
\begin{gather}
p_T = {\bf w_o} H_T,~~~ p_S = {\bf w_o} H_S 
\end{gather}

\begin{table*}[htp]
\small
\begin{center}
\begin{tabular}{p{6cm}  c c c c c c}
\toprule
\multirow{2}*{\bf System} & \multicolumn{2}{c}{\bf RACE} & \multicolumn{2}{c}{\bf DREAM} & \multicolumn{2}{c}{\bf C$^3$} \\
& \bf Dev & \bf Test & \bf Dev & \bf Test & \bf Dev & \bf Test \\
\midrule
ALBERT-base \citep{zhu-etal-2020} & - & - & 64.5 & 64.4 & - & - \\
ALBERT-base + DUMA \citep{zhu-etal-2020} & - & - & 67.1 & 67.6 & - & - \\
ALBERT-base + TB-DUMA \citep{zhu-etal-2020} & - & - & 67.8 & 67.2 & - & - \\
ALBERT-base \citep{clue} & - & - & - & - & 60.4 & 59.6 \\
ALBERT-base \citep{Lan2019ALBERT} & - & 66.8 & - & - & - & - \\
\midrule
ALBERT-base (re-run) & 71.1 \tiny(70.8) & 69.7 \tiny(69.1) & 68.0 \tiny(67.1) & 66.7 \tiny(66.1) & 60.3 \tiny(60.1) & 60.0 \tiny(59.5) \\
ALBERT-base + RDG (constant) & 70.9 \tiny(70.6) & 70.1 \tiny(69.6) & \bf 69.1 \tiny(67.8) & \bf 68.4 \tiny(67.1)  & \bf 60.9 \tiny(60.2) & \bf 60.6 \tiny(60.2) \\
ALBERT-base + RDG (recursive) & \bf 72.3 \tiny(71.4) & \bf 71.4 \tiny(70.3) & 68.1 \tiny(67.4) & 68.2 \tiny(66.9) & 60.5 \tiny(59.8) & 60.1 \tiny(59.7) \\
\bottomrule
\end{tabular}
\caption{\label{result-baseline} Experimental results on RACE, DREAM, and C$^3$. The average score is depicted in brackets.} 
\end{center}
\end{table*}

\subsection{Training Objective}
The training objective is constructed in three parts.
\begin{gather}
\mathcal{L} = \mathcal{L}_{ce} + \mathcal{L}_{kd} + \mathcal{L}_{cos} 
\end{gather}
First, we will maximize the output for both teacher and student using standard cross-entropy loss, as shown below, where $p'_T$ and $p'_S$ is the softmaxed values respectively.
\begin{gather}
\mathcal{L}_{ce} = -\frac{1}{2N} \sum\limits_{i=1}^{N}(y_T\log(p'_T) + y_S\log(p'_S) )
\end{gather}

Compared to the teacher, the student will use a filtered input for each transformer layer and thus will learn much slower. 
In this case, we apply an additional knowledge distillation (KD) loss, specifically MSE loss in this paper, between the logits of the teacher and student model to accelerate the learning, as shown below, where the $T$ is the temperature argument.
\begin{gather}
\mathcal{L}_{mse} = \frac{1}{N} \sum\limits_{i=1}^{N}(\frac{p_T}{T} - \frac{p_S}{T})^2
\end{gather}

By only using the first two losses are not enough, as the model will only optimize for the final accuracy and could result in all-one gating values for any input, which will be identical to the teacher model and will not be informative in explainability (see Section \ref{sec-ablation} for further discussion).
In this context, at last, we calculate the cosine similarity between the input mask $I \in\mathbb{R}^{\mathcal \\l}$ and the final gating value $R_L$ to encourage the model to achieve better scores (cross-entropy and KD loss) with less input information (cosine similarity loss).
\begin{gather}
\mathcal{L}_{cos} = \frac{1}{N} \sum\limits_{i=1}^{N}cosine(I, R_L)
\end{gather}

\section{Experiments}
\subsection{Experimental Setups}
We evaluate our approach on three multiple-choice machine reading comprehension datasets, including RACE \citep{lai-etal-2017}, DREAM \citep{sun-etal-2019-dream}, and C$^3$ \citep{sun-etal-2020-c3}, where the last one is a Chinese MRC dataset. 
Unlike the span-extraction machine reading comprehension (such as well-known SQuAD \citep{rajpurkar-etal-2016}), where the answer is an exact span in the passage, the multiple-choice machine reading comprehension requires to select correct answer among several candidates, which is not directly picked from the original passage and is suitable to test the explainability of our model.
The statistics of these datasets are shown in Table \ref{dataset-stat}.
Note that, for RACE and DREAM datasets, there are always four candidate answers.
However, for C$^3$ dataset, the number of the candidate answers ranges from 2 to 4, and we manually add dummy candidates to those samples that have less than four candidates.
As indicated in the previous section, we use the recursive history for the RACE and the constant history for the other datasets.

\begin{table}[ht]
\small
\begin{center}
\begin{tabular}{p{3cm} rrr}
\toprule
& \bf RACE & \bf DREAM & \bf C$^3$ \\
\midrule
Train \# & 87,866 & 6,116 & 8,023 \\
Dev \# & 4,887 & 2,040 & 2,674 \\
Test \# & 4,934 & 2,041 & 2,672 \\
Choices \# per question & 4 & 4 & 2 \~{} 4 \\
\midrule
Gating decay rate & 0.9 & 0.8 & 0.8 \\
Conv kernel size & 3 & 3 & 5 \\
Initial learning rate & 2e-5 & 1e-5 & 2e-5 \\
Training steps & 1,200 & 1,000 & 2,000 \\
\bottomrule
\end{tabular}
\caption{\label{dataset-stat} Data statistics and hyper-parameter settings.}
\end{center}
\end{table}

We also list the most important hyper-parameter settings for each task in Table \ref{dataset-stat}. 
Following original ALBERT implementation, for all tasks, we use a maximum sequence length of 512, and classifier dropout of 0.1, and ALBERT dropout of 0. 
The temperature of the MSE loss is set to 8, which is a common choice in knowledge distillation settings, as suggested in \citet{textbrewer-acl2020-demo}.
We use \textsc{Adam} \citep{kingma2014adam} with weight decay optimizer with a warm-up rate of 0.1. The training batch size is set to 32.
We use ALBERT-base-v2 checkpoint for English and ALBERT-base-zh checkpoint for Chinese to initialize all transformer layers.\footnote{https://github.com/google-research/albert} 
To ensure the stability of the results, for each experiment, we perform ten training trials with different random seeds and report the maximum and average scores.
Our implementation is based on the TensorFlow \citep{abadi2016tensorflow} version of {\tt run\_race.py} provided by ALBERT. 
All models are trained on Cloud TPU v2 with 64GB HBM.

\begin{table*}[ht]
\small
\begin{center}
\begin{tabular}{p{2cm} p{8cm} r r}
\toprule
\bf Score & \bf Description & \bf Attention & \bf RDG \\
\midrule
5-Excellent & Sufficient explanations, very few redundant text segments & 1\% & 46\% \\
4-Acceptable & Sufficient explanations, contains a few redundant text segments but not exceed the core explanations & 17\% & 34\% \\
3-Marginal & Almost sufficient explanations, but requires the further information in the original passage & 21\% & 10\% \\
2-Trivial & Only a few explanations with redundant text segments that exceed the core explanations & 29\% & 8\% \\
1-Wrong & No informative explanations provided, or completely unreadable & 32\% & 2\% \\
\midrule 
\multicolumn{2}{l}{\bf Weighted Average Score}  & 2.26 & \bf 4.14 \\
\bottomrule
\end{tabular}
\caption{\label{human-score} Human evaluation on answer explainability over 100 correctly predicted samples in the RACE test set. Importance: Usability $>$ Redundancy $>$ Readability}
\end{center}
\end{table*}

\subsection{Results}
We will evaluate our model with respect to the objective benchmarks to see if the teacher model could benefit from a student that is good at extracting explanations.
The overall experimental results are shown in Table \ref{result-baseline}.
To ensure a fair comparison, we re-run the baseline system and yields better results (except for the development set of C$^3$) than previous state-of-the-art systems on the ALBERT-base model.
On top of our baseline system, as we can see that, our model yields stable and consistent improvements in all three datasets, indicating that by simultaneously learning the teacher and student model could improve the overall prediction accuracy.
Compared to the best scores in RACE, the accuracy of the student output is 67.8/66.8, which is lower than the teacher system, and thus we stick to use the teacher system to make the final prediction.\footnote{We also tried to combine the output of both teacher and student system, but did not find additional gains.}
As the main body system, i.e., ALBERT model, is shared between the teacher and student, the learning of the student will also `regularize' the teacher, and thus the final prediction of the teacher system will improve against the baseline.
Also we have found that recursive history mechanism (Eq. 6a) is efficient for the larger dataset (i.e., RACE), which demonstrate that the learning for the history requires more training data.

\subsection{Ablation Study}\label{sec-ablation}
We ablate critical components in our RDG system to separately evaluate their contributions to the whole system.
Specifically, we ablate the RDG system with recursive history on RACE dataset.
Note that, unlike most of the previous works that only present a single-run in ablations, in this paper, all ablation experiments are also conducted ten times, and we report their average scores to ensure the reliability of the results, as shown in Table \ref{ablation}. 

\begin{table}[htbp]
\small
\begin{center}
\begin{tabular}{p{4cm} c c c}
\toprule
& \bf Dev & \bf Test & \bf Avg \\
\midrule
ALBERT-base+RDG & 71.4 & 70.3 & 70.9 \\
~~~w/o cosine loss & 71.3 & 70.2 & 70.8 \\
~~~w/o KD loss & 71.3 & 69.7 & 70.5 \\
~~~w/o conv smoothing & 71.0 & 70.0 & 70.5 \\
~~~w/o historical gating & 70.9 & 69.8 & 70.4 \\
ALBERT-base & 70.8 & 69.1 & 70.0 \\
\bottomrule
\end{tabular}
\caption{\label{ablation} Results of model ablations on RACE.}
\end{center}
\end{table}

Historical gating and convolutional smoothing are the most important components in our RDG mechanism, demonstrating that it is necessary to build a well-designed architecture for the gating mechanism to effectively filter the information.
The KD loss seems to also be effective in our system, indicating that the KD loss will accelerate the learning of the student system. As a result, the accuracy of student output will rise from 60.8/60.3 to 67.8/66.8 (ALBERT-base+RDG, as indicated in the previous section), which also verify this assumption.
Eliminating the cosine loss seems to be less effective in improving the final performance.
After careful examination, when removing the cosine loss, the gating values for most of the words will be one and RDG loses its explainability, as the student will learn to let all information pass to the transformer layers to maximize the objective performance instead of pointing out the key components.
This indicates that though the cosine loss does not improve the overall accuracy, it is essential to achieve better explainability in our RDG system.

\begin{table*}[ht]
\tiny
\begin{center}
\begin{tabular}{p{3cm} p{4cm} p{3.5cm} p{3cm}}
\toprule
\multicolumn{4}{p{15cm}}{{\bf Passage 1}: water is very important to living things. without water there can be no life on the earth. all animals and plants need water. man also needs water. we need water to drink, to cook our food and to clean ourselves. water is needed in offices, factories and schools. water is needed everywhere. there is water in seas, rivers and lakes. water is found almost everywhere. even in the desert part of the world, there is some water in the air. you can not see or feel it when it is a part of the air. the water in the seas, rivers and lakes is a liquid, the water in the air is a gas, and we call it water vapour . clouds are made of water. they may be made of very small drops of water. they may also be made of snow crystals . snow crystals are very very small crystals of ice. ice is frozen water. it is a solid. there can be snow and ice everywhere in winter. water may be a solid or a liquid or a gas. when it is a solid, it may be as hard as a stone. when it is a liquid, you can drink it. when it is a gas, you can not see or feel it.} \\
\midrule
\bf  Question & \bf Candidates & \bf Attention & \bf RDG \\
\midrule
where can we find water? & \makecell[lt]{A: we can find water when it turns into vapor \\ B: water is only in seas and rivers \\ C: we can see water in deserts here and there \\ {\bf D: water can be found almost everywhere}} & {water is very important to living things. without water. can be no life on...is needed everywhere. there... feel it.} &  {... rivers and lakes. water is found almost everywhere...} \\
\midrule
clouds are made of & \makecell[lt]{A: seas, rivers and lakes \\ B: blocks of ice \\ {\bf C: very small drops of water or snow crystals} \\ D: solid, liquid and gas} & {water is very important to living things... you can not see or... winter. water may be... you can drink it...feel it.} & {...they may be made of very small drops of water. they may also be made of snow...} \\
\midrule\midrule
\multicolumn{4}{p{15cm}}{{\bf Passage 2}:  is it important to have breakfast every day? a short time ago, a test was given in the united states. people of different ages, from 12 to 83, were asked to have a test. during the test, these people were given all kinds of breakfast, and sometimes they got no breakfast at all. scientists wanted to see how well their bodies worked after eating different kinds of breakfast. the results show that if a person eats a right breakfast, he or she will work better than if he or she has no breakfast. if a student has fruit, eggs, bread and milk before going to school, he or she will learn more quickly and listen more carefully in class. some people think it will help you lose weight if you have no breakfast. but the result is opposite to what they think. this is because people become so hungry at noon that they eat too much for lunch. they will gain weight instead of losing it.} \\
\midrule
\bf  Question & \bf Candidates & \bf Attention & \bf RDG \\
\midrule
what were the people given during the test? & \makecell[lt]{A: no breakfast at all. \\ B: very rich breakfast. \\ C: little food for breakfast. \\ {\bf D: different foods or sometimes none.}} & {is it important ... every day? a short ... a test.during the test, these people were given all kinds of... losing it.} &  {... given all kinds of breakfast, and sometimes they got no breakfast at all...} \\
\midrule
what do the results show? & \makecell[lt]{{\bf A: breakfast has affected on work and studies.} \\ B: breakfast has little to do with a person's work. \\ C: will work better if he only has fruit and milk. \\ D: girl students should have less for breakfast.} & {is it important... every day? a short... of breakfast. the results show that if... no breakfast.if a... result is opposite to what... losing it.} & {...show that if a person eats a right breakfast, he or she will work better than if... will learn more quickly and...
}\\
\midrule\midrule
\multicolumn{4}{p{15cm}}{{\bf Passage 3}: traditionally, women have fallen behind men in adoption of internet technologies, but a study released yesterday by the pew internet \& american life project found that women under age 65 now get ahead of men in internet usage, though only by a few percentage points. but the survey also noted that the between women and men on the web is even greater among the 18 to 29 age group and african americans. the report, "how women and men use the internet", examined use by both sexes, looking at what men and women are doing online as well as their rate of adopting new web-based technologies. the report found that 86 percent of women aged 18 to 29 were online, compared with 80 percent of men in the same age group. among african americans, 60 percent of women are online, compared with 50 percent of men. in other age groups, the disparity is only slight, with women outpacing men by 3 percentage points. however, among the older group, those age 65 and older, 34 percent of men are online, compared with 21 percent of women. men tend to use the web for information and entertainment, getting sports scores and stock quotes and downloading music, while women tend to be heavier users of mapping and direction services, and communication services such as email. }  \\
\midrule
\bf  Question & \bf Candidates & \bf Attention & \bf RDG \\
\midrule
among african americans, women & \makecell[lt]{A: equal men in the use of the internet . \\B: use the internet less than men . \\{\bf C: use the internet more than men .} \\D: use the internet better than men .} & {... age 65 now get ahead of men..., though only by a... 86 percent of women aged. to 29 were online,... women. men tend ... e-mail.} &  {..., 60 percent of women are online, compared with 50 percent of men ...
} \\
\bottomrule
\end{tabular}
\caption{\label{case-study} Explainability comparisons of the attention mechanism and our RDG system on correctly predicted samples in the RACE test set. The `...' indicate the omitted passage words. Ground truths are depicted in boldface.}
\end{center}
\end{table*}

\section{Explainability Evaluation}
\subsection{Human Evaluation}
Though our RDG system yields better performance in the objective benchmarks, our primary goal is to improve the explainability of the MRC systems.
Unlike the explanation system with supervised learning, which could be evaluated through the comparisons between the prediction and ground truth, in this paper, we carried out human evaluations to demonstrate the explainability of the proposed method, as no ground truths are available.
We compared our RDG approach with the traditional in-built multi-head attention mechanism to evaluate which one provides better clues for the answer.
We perform the following steps for each system.
Note that we extract these explanations with respect to the ground truth candidate.
\begin{itemize}[leftmargin=*]
	\item {\bf Attention}: as in-built multi-head attention in BERT-like language models are in 4D (batch, num\_att\_heads, seq\_length, seq\_length\_softmax), we perform the average operation through attention head dimension and sequence length dimension to get a flattened 2D attention score. We only print out attention scores in the passage.\footnote{We also tried to examine all the attention heads (12*12=144), but did not find a specific head for this purpose across different samples.}
	\item {\bf RDG}: we directly extract the final gating values to represent the importance of each input word.
\end{itemize}
After obtaining these scores, in descending order, we select the top 10 words and their contextual words ($\pm 2$ positions) to demonstrate the explainability.
We randomly select 100 questions among the correctly predicted samples in the RACE test set.
The annotators are instructed to first read the question and candidate answers, and then read the explanations of two systems.
When the explanations are not readable, the annotators are allowed to see the original passage to assess their usability.

According to the quality of the explanations for the correct answer, the annotator will grade two systems, ranging from 1-wrong to 5-excellent. 
The results of the human evaluation, as well as the standard for each score, are listed in Table \ref{human-score}.
As we can see that the proposed RDG system could substantially improve the explainability over traditional in-built attention mechanism, with an average score of 4.14, which is `acceptable'.
The RDG system will extract much more meaningful explanations with 80\% for `acceptable` and `excellent' grades, while the traditional attention mechanism only takes 18\%.

We find the traditional attention mechanism is not that straightforward for the human to explicitly know the explanations for the answer, as it is hard to find a `conclusive' head for the overall attention in the multi-head mechanism.
Nonetheless, as the multi-head attention mechanism will implicitly learn the semantic, dependency, and other linguistic knowledge in each head, it is interesting to see if we could utilize them to generate much more meaningful explanations for humans, and we will leave it for future work.

\subsection{Case Analysis}
We select several cases to analyze the linguistic aspect of the generated explanations, as shown in Table \ref{case-study}.
As we can see that the explanations generated by the RDG system substantially outperform the traditional attention mechanism in terms of the usability, readability, and variety, where we conclude our observations as follows.
\begin{itemize}[leftmargin=*]
	\item {\bf Usability}: The most important aspect is the usability of the explanations. Our RDG system could give relatively informative explanations to the answer in terms of the keywords. For example, in passage 1 question 1, both explanations could identify the keyword `everywhere' in the option D. However, from the original passage, we can know that the word `everywhere' in the explanation of the attention indicate `water is needed everywhere', which is not accordance with the question, while our RDG system correctly identifies the explanation `water is found almost everywhere'. 
	In another case in the passage 2 question 2, our RDG system could also extract clues in discontinued span, where `work and studies' in the candidate A mapped to two discontinued span `work better' and `learn more quickly' in the original passage.
	\item {\bf Readability}: The explanations generated by attention contain more text segments than the RDG system, and some of them are hard for the human to understand. On the contrary, the RDG system could generate more fluent explanations. As the explanations are identified not only by the top-$k$ words but also their context, it could be inferred that our model could generate more concentrated explanations rather than scattered ones.
	\item {\bf Variety}: The explanations generated by attention show a strong pattern that the most front and end words are always be chosen, which is the closest term to {\tt [CLS]} and {\tt [SEP]} tokens in ALBERT (or other BERT-like models). For example, in passage 1, all explanations generated by the attention mechanism include the first sentence `water is very important to living things'. This observation is similar to the previous study by \citet{kovaleva-etal-2019-revealing}, that a majority of attention maps show strong attention to the {\tt [CLS]} and {\tt [SEP]} or the token itself. However, our RDG system will generate varied explanations according to different questions instead of meaningless special tokens or repetitive expressions in the human view.
\end{itemize}

It is worth mentioning that not all cases are explainable by `text matching' or `semantic matching'.
In passage 3, there is no obvious sentences that matches the correct candidate answer `use the internet more than men'.
However, our RDG system could identify the correct clues for the answer, mentioning the higher percentage of women than men in the internet usage. 
While the attention mechanism failed to identify the correct explanations.

\subsection{Negative Sample Analysis}
In the previous sections, we show and analyze the explanations for the correctly predicted samples. 
However, though the explanations for the wrongly predicted sample is less informative to the human, we also wonder if we could find clues for these samples.
We also manually select 100 wrongly predicted samples and see whether the explanations corresponding to the ground truth candidate answer are informative to evaluate their potentials in choosing a better candidate.
The results show that the attention system only has 18\% samples gives reasonable explanations for the correct answer, while our RDG system has a high percentage of 56\%, indicating that our system has a better correlation between the ground truth and its explanations in the passage.

\section{Conclusion}\label{conclusion}
In this paper, we aim to improve the explainability for the machine reading comprehension task, which is different from most of the previous works that were only striving for better objective evaluation scores. 
To achieve this goal, we propose a novel mechanism called Recursive Dynamic Gating (RDG) to gradually refine the amount of the input information in each layer of the pre-trained language model.
Also, we propose an attention smoothing technique that will increase the accuracy of the RDG mechanism.
Experimental results on three multiple-choice machine reading comprehension datasets show that the proposed RDG mechanism could not only improve the objective evaluation scores, but also show an advantage over the traditional attention mechanism in explainability.

As the explainability will become an important topic in future research for machine reading comprehension as well as more generalized natural language processing tasks, in the future, we are going to further utilize the explanation segments as a source to generate more fluent and comprehensive explanations.

\bibliography{emnlp2020}
\bibliographystyle{acl_natbib}

\end{CJK*}
\end{document}